# Language-guided Medical Image Segmentation with Target-informed Multi-level Contrastive Alignments

Mingjian Li [a, b, †], Mingyuan Meng [a, b, c, †], Shuchang Ye [a], Mingye Zou [c], Michael Fulham [a, d], Lei Bi [a, b, *], and Jinman Kim [a, *]

[a] School of Computer Science, The University of Sydney, Sydney, Australia.
[b] Institute of Translational Medicine, Shanghai Jiao Tong University, Shanghai, China.
[c] Zhongguancun Academy & Zhongguancun Institute of Artificial Intelligence, Beijing, China.
[d] Department of Molecular Imaging, Royal Prince Alfred Hospital, Sydney, Australia.

[†] M. Li and M. Meng contributed equally to this work.
[*] Corresponding authors: lei.bi@sjtu.edu.cn (L. Bi), and jinman.kim@sydney.edu.au (J. Kim)

**Abstract —** Medical image segmentation is a fundamental task in numerous medical applications. Recently, language-guided segmentation has shown promise in medical scenarios where textual clinical reports are readily available as semantic guidance. Clinical reports contain diagnostic information provided by clinicians, which can provide auxiliary textual semantics to guide segmentation. However, existing language-guided segmentation methods neglect the inherent pattern gaps between image and text modalities, resulting in sub-optimal visual-language integration. Contrastive learning is a well-recognized approach to align image-text patterns, but it has not been optimized for medical image segmentation, where clinically meaningful semantics are often concentrated in localized target regions rather than the entire image. In this study, we propose TMCA, a Target-informed Multi-level Contrastive Alignment framework to bridge image-text pattern gaps for medical language-guided segmentation. The core innovation is to reformulate image-text contrastive alignment from conventional instance-level matching to segmentation-oriented semantic matching, where image-text samples are aligned according to their segmentation targets rather than merely whether they come from the same patient. Specifically, TMCA enables target-informed image-text alignments and fine-grained textual guidance by introducing: (i) a target-sensitive semantic distance module that utilizes target information for more granular image-text alignment modeling, (ii) a multi-level contrastive alignment strategy that directs fine-grained textual guidance to multi-scale image details, and (iii) a language-guided target enhancement module that reinforces attention to critical regions based on the aligned image-text patterns. Extensive experiments on four public benchmarks, involving three medical imaging modalities with clinical reports, show that TMCA enabled superior performance over state-of-the-art language-guided medical segmentation methods.

## 1. Introduction

Medical image segmentation is a fundamental task in a wide range of medical applications, including computer-aided diagnosis, radiation therapy planning, and surgical navigation (Chen et al., 2022). The aim of medical image segmentation is to identify regions of interest (ROIs) in medical images, e.g., identifying and quantifying tumor ROIs in cancer patient scans is a key factor in determining disease progression and the effectiveness of treatment (Jiang et al., 2023). In recent years, deep learning-based methods have achieved notable success for automatic medical image segmentation (Liu et al., 2025; Wang et al., 2022b; Yu et al., 2025). While early segmentation methods typically focused on exploiting image information, leveraging textual clinical reports as complementary information for segmentation has gained wide attention for its potential to transcend the performance bounds of image-only segmentation methods. Clinical reports, formulated by clinical experts, are readily available with the images and can provide valuable semantic information, including diagnostic imaging findings and interpretations. Recently, language-guided segmentation methods, which leverage auxiliary textual semantics in clinical reports to guide segmentation, have emerged to boost segmentation performance (Li et al., 2023; Lee et al., 2023; Tomar et al., 2022; Zhong et al., 2023).

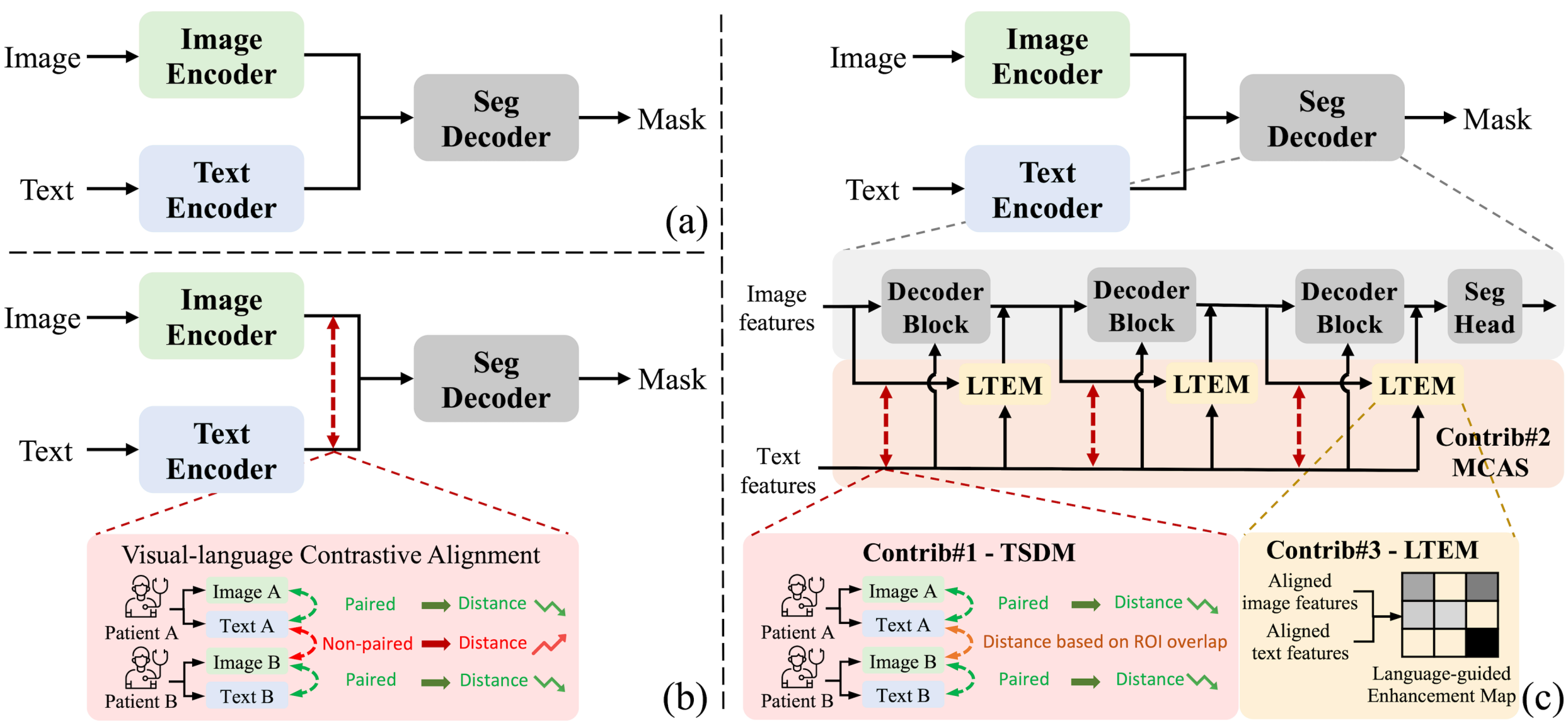

Figure 1. Illustration of (a) language-guided segmentation that directly integrates image and text features, (b) language-guided segmentation enhanced by visual-language contrastive alignments, and (c) the proposed Target-informed Multi-level Contrastive Alignment framework (TMCA) for language-guided segmentation, featuring three technical contributions: Target-sensitive Semantic Distance Module (TSDM, Contrib#1), Multi-level Contrastive Alignment Strategy (MCAS, Contrib#2), and Language-guided Target Enhancement Module (LTEM, Contrib#3).

Language-guided medical image segmentation methods typically extract image and text features using separate encoders and integrate them via direct visual-language feature fusion (Figure 1a), exploiting textual semantic guidance with visual contexts (Lee et al., 2023; Zhong et al., 2023). However, despite the improved segmentation performance, these methods neglect the pattern gaps between image and text modalities during feature fusion, resulting in sub-optimal information integration and less-effective textual guidance. This is because image and text modalities are inherently encoded in different patterns: images are continuous and rich in details, whereas text is discrete and concise with a focus on high-level concepts. Moreover, the encoders used for extracting image and text features are fundamentally different, further enlarging the pattern gaps between the extracted image and text features.

Vision-language contrastive learning can bridge image-text pattern gaps by associating the semantics of paired image-text features through contrastive alignments (Zhang et al., 2022; Zhao et al., 2023). In language-guided segmentation, standard contrastive alignment can be used to align the image and text features before feature fusion (Figure 1b), which attempts to minimize the semantic distance between paired image-text features (from the same patient) while maximizing the semantic distance between non-paired features (from different patients). Unfortunately, this straightforward adoption does not optimize contrastive alignment for medical image segmentation and, therefore, leaves considerable research gaps. Existing contrastive alignments typically align global image-text semantics holistically, favoring vision tasks related to global semantics (such as image classification); however, segmentation aims to delineate local ROIs in medical images, making it a localized task instead of characterizing the entire image. This conflict incurs limitations specifically in two ways: (i) First, current contrastive alignment techniques have no awareness of ROI target information and assume exact correspondence between paired image-text features (i.e., assume that they are from the same patient) (Zhang et al., 2022), facing challenges when aligning image-text pairs from different patients that, while not identical in holistically, have similar ROIs. Consequently, there may be under-alignment between image-text features that are semantically related but derived from different patients, and this might disrupt the feature alignment process and lead to poor feature fusion. (ii) Second, medical image segmentation, as a pixel-wise prediction task, relies primarily on subtle low-level image details to characterize the underlying disease/targets and identify the ROI boundaries (Meng et al., 2023). However, current contrastive alignment techniques typically align high-level semantics only at the deepest level of encoders, thereby preventing the delivery of fine-grained textual guidance directly on subtle medical image details.

In this study, we propose TMCA, a novel Target-informed Multi-level Contrastive Alignment framework particularly designed to bridge the image-text pattern gaps for language-guided medical image segmentation. As illustrated in Figure 1c, our TMCA incorporates ROI target information into image-text alignment modeling and directs fine-grained textual guidance to crucial medical image details, featuring three technical contributions as follows:

• Target-sensitive Semantic Distance Module (TSDM) calculates semantic distance based on the intersection ratios of ROIs, thus enabling alignments between image-text features that describe similar ROIs but differ in less-important backgrounds.

• Multi-level Contrastive Alignment Strategy (MCAS) performs contrastive alignments at multiple feature pyramid levels, hence delivering fine-grained textual guidance on multi-scale image features containing rich image details.

• Language-guided Target Enhancement Module (LTEM) reinforces attention to critical image regions by explicitly associating the aligned image-text features with target-related image sub-regions.

Extensive experiments were conducted on four public benchmark datasets across three medical image modalities, including X-ray (QaTa-COV19 (Degerli et al., 2022)), computed tomography (MosMedData (Morozov et al., 2020)), and endoscopic images (Kvasir-SEG (Jha et al., 2019) and Bkai-polyp (Phan et al., 2021)). The experimental results showed that our TMCA consistently outperformed state-of-the-art language-guided medical image segmentation methods across all four evaluation datasets, underscoring its generalizability and effectiveness.

## 2. Related Work

### 2.1. Deep Learning for Medical Image Segmentation

Deep learning has made substantial advances in medical image segmentation. The U-Net (Ronneberger et al., 2015) using a convolution neural network (CNN) has demonstrated outstanding performance for biomedical image segmentation, whose U-shaped architecture is composed of a contracting encoder for contextual understanding and a symmetric expanding decoder for ROI delineation. This architecture has been shown to be instrumental in learning multi-scale feature representations for precise segmentation. U-Net++ (Zhou et al., 2018) then re-vamped skip connections to mitigate the semantic gaps between the encoder and decoder. Attention U-Net (Oktay et al., 2018) introduced attention gates to suppress non-target backgrounds. Swin UNETR (Hatamizadeh et al., 2021) introduced transformers with shifted windows to capture long-range contexts across image sub-regions. Meanwhile, Isensee et al. (2021) proposed nnU-Net to automate the optimal configuration process of U-Net. Ma et al. (2024; 2025) proposed a foundation segmentation model, MedSAM, trained on large-scale medical data and demonstrated great generalizability across datasets. Furthermore, segmentation methods using multi-modal medical images are gaining prominence over single-modal methods. Kamnitsas et al. (2017) proposed a segmentation model that combines multiple CNNs, each trained on different medical imaging modalities. Guo et al. (2019) investigated three distinct strategies for integrating multi-modal medical images, noting that fusing multi-modal information within the network contributed to superior performance when compared to outcome-level fusion (e.g., majority voting). Although deep learning has made impressive advances in medical image segmentation, its success is greatly reliant on the availability of large-scale labeled training data, which are usually tedious and time-consuming to annotate and difficult to acquire due to various ethical constraints (Shen et al., 2017).

### 2.2. Language-guided Medical Image Segmentation

To exploit readily available textual clinical reports to improve segmentation accuracy, language-guided segmentation methods have emerged and shown promising performance (Xue et al., 2026; Zeng et al., 2026). For instance, Tomar et al. (2022) extracted textual semantics via 'byte-pair' encoding and employed it as soft channel attention to reinforce critical image features. Li et al.

(2023) proposed LViT that integrated image and text features using a U-shaped CNN alongside a U-shaped Visual Transformer (ViT) (Dosovitskiy et al., 2021). These methods extracted text features via simple vectorization operations, incurring difficulties in capturing rich textual semantics. To mitigate this, Lee et al. (2023) employed the text encoder of CLIP (Radford et al., 2021), which was pretrained on a large language corpus and had proven effective for various textual tasks for feature extraction, and then integrated the extracted text features with image features via cross-attention. To further accommodate various medical concepts and specialized knowledge in clinical reports, Zhong et al. (2023) employed a medical domain-specific text encoder, CXR-BERT (Boecking et al., 2022), to extract text features encoded with rich semantics and then fuse them with image features. Hu et al. (2024) proposed to extract image features from the pretrained Segment Anything Model (SAM) (Kirillov et al., 2023) and integrated them with text features extracted from the pretrained BERT (Devlin et al., 2019) model for segmentation. Recently, Huang et al. (2025) proposed a cross-modal conditioned reconstruction model (RecLMIS) for medical language-guided segmentation, which explicitly captured image-text interactions by conditioned vision-language reconstruction. These methods focused on enhancing segmentation accuracy with textual semantics and required textual reports as input during both training and inference. To alleviate the reliance on textual input, Ye et al. (2024; 2025) performed joint report generation or leveraged prototype learning to generate textual semantics for inference without necessitating textual reports as input. Unfortunately, these methods compromised the segmentation accuracy compared to common language-guided segmentation methods that require textual reports during inference.

Most language-guided medical image segmentation methods focused on feature extraction and fusion strategies to establish image-text semantic interactions, but neglected the inherent pattern gaps between the two distinct image and text modalities. Only a few studies attempted to bridge the pattern gaps in language-guided segmentation. For example, Pan et al. (2026) adopted evidential learning to bridge the pattern gaps by measuring the modality gap based on opinion aggregation; Zhou et al. (2025) adopted contrastive learning to holistically align the features of image and text encoders via a multi-stage cross-modality contrastive loss. These methods alleviate the image-text pattern gaps via established evidential/contrastive learning techniques; however, they failed to tailor the techniques for language-guided segmentation tasks, e.g., incorporating ROI target information into alignment modeling.

### 2.3. Contrastive Alignment

Vision-language contrastive learning is a widely used technique that learns the general image representations from language supervision, where text acts as "pseudo instance-level classification labels". CLIP (Radford et al., 2021) is a pioneering work that involved over 400 million image-text pairs. Zhang et al. (2022) introduced contrastive learning to the medical imaging domain, which minimized the distance between paired image-text features from the same patient while maximizing the distance between non-paired features from different patients. Also, Huang et al. (2021) proposed joint global-local contrastive representation learning, Wang et al. (2022a) further explored multi-granularity feature alignment, and Li et al. (2025) proposed to model image-text inter-match relations. Once pretrained, the image encoder can be applied to various downstream tasks using task-specific decoders, such as classification, detection, and segmentation (Zhao et al., 2023). However, this pretraining-based approach cannot explicitly extract and leverage textual semantics for inference. Therefore, in this study, we explored a different approach toward contrastive image-text alignment tailored for segmentation: the contrastive alignments are deeply embedded into the segmentation decoder, so that the whole encoder-decoder framework can jointly learn and collectively infer for segmentation with better utilization and integration of image-text features.

In addition, medical images and reports of different patients can have large semantic similarities, e.g., describing similar diseases. Recently, Wang et al. (2022c) proposed to measure alignment distance based on disease classification labels, and Liu et al. (2023) also proposed to use text feature similarities for semantically-aware alignment distance measurement. However, these methods were not applicable to segmentation tasks: for example, the disease labels could be the same for different patients, while the ROI locations and shapes vary substantially. To the best of our knowledge, our TMCA is the first contrastive alignment framework tailored for language-guided segmentation, focusing on localized, target-sensitive alignment for clinically-critical ROIs.

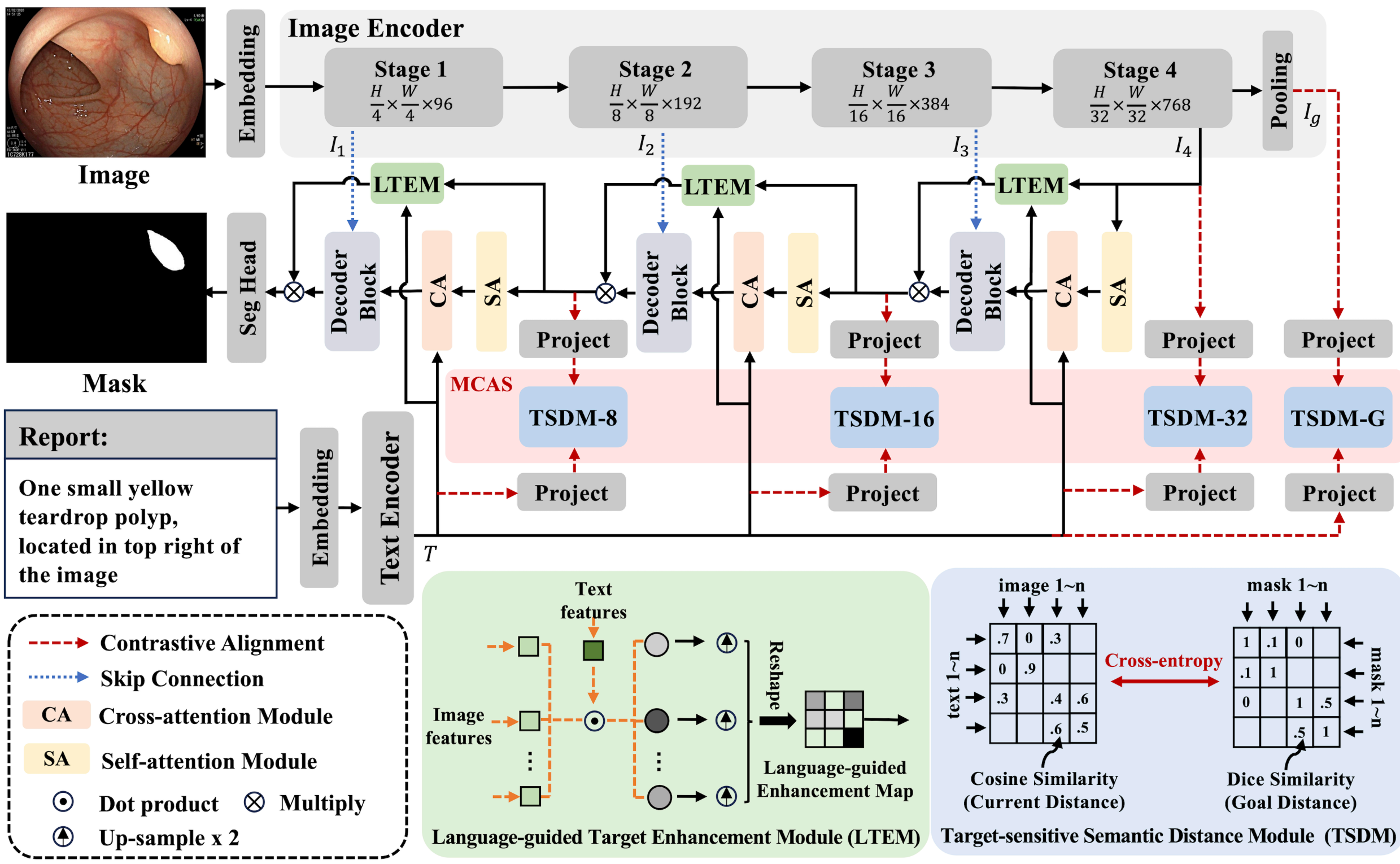

Figure 2. Overall architecture of our TMCA for language-guided medical image segmentation, containing image/text encoders and a segmentation decoder with target-informed multi-level contrastive alignments.

## 3. Method

In language-guided medical image segmentation, the central challenge is not only to fuse visual and textual features, but to align them according to target-relevant semantics. Existing multimodal alignment strategies are mainly designed for patient-level matching, image-text retrieval, or classification-oriented semantics. By contrast, TMCA is built on a segmentation-specific principle: cross-modal similarity should be defined by target-informed semantic consistency, propagated across hierarchical visual representations, and finally translated back into localization-sensitive feature enhancement. Under this principle, TSDM, MCAS, and LTEM jointly reduce the modality gap between medical images and clinical text in a manner tailored to segmentation.

The overall architecture of our TMCA is illustrated in Figure 2, which consists of image/text encoders (detailed in Section 3.1) and a segmentation decoder equipped with target-informed multi-level contrastive alignments (detailed in Section 3.2). The decoder part contains our three technical contributions: TSDM (Section 3.2.1), MCAS (Section 3.2.2), and LTEM (Section 3.2.3).

### 3.1. Image and Text Encoders

ConvNeXt-Tiny (Liu et al., 2022) was used as the image encoder following Zhong et al. (2023), where multi-scale image features were extracted from four stages of the image encoder, denoted as $[I_1, I_2, I_3, I_4]$. We also extract global image features using a global pooling layer after the last stage of the image encoder, which is denoted as $I_g$.

CXR-BERT (Boecking et al., 2022), a medical domain-specific language model, was employed as the text encoder. CXR-BERT was pretrained on large-scale medical reports and has shown great capability in textual semantics extraction. We extract text features from the last three layers of CXR-BERT, and then average them to derive the final text features $T$.

### 3.2. Segmentation Decoder with Target-informed Multi-level Contrastive Alignments

The segmentation decoder progressively integrates features from the image and text encoders, which consists of three successive decoder blocks followed by a segmentation head that projects the final feature maps into segmentation masks. The segmentation loss is defined as the sum of Dice Loss and Cross-Entropy Loss.

Before each decoder block, the input image and text features are semantically aligned via contrastive alignments based on our Target-sensitive Semantic Distance Module (TSDM, detailed in Section 3.2.1). In addition to the TSDM modules used before the three decoder blocks, another TSDM module is used to align the global semantics between $I_g$ and $T$, totaling four levels of contrastive alignments in our Multi-level Contrastive Alignment Strategy (MCAS, detailed in Section 3.2.2).

Within each decoder block, the TSDM-aligned image features with a positional embedding are fed into a multi-head self-attention (SA) module to capture long-range intra-modal relations, followed by a multi-head cross-attention (CA) module to integrate the SA-output image features with the TSDM-aligned text features (image features serve as $Q$ and text features serve as $K$/$V$). Both the SA and CA modules have shortcut connections between their inputs and outputs. After that is a standard U-Net upsample block with a skip connection from the image encoder; specifically, a deconvolution layer upsampled the feature maps by a factor of 2, followed by two convolution layers to integrate the image features propagated from the skip connection.

After each decoder block, a Language-guided Target Enhancement Module (LTEM, detailed in Section 3.2.3) is used to exploit the aligned image-text semantics to guide attention to critical image regions. The output features of the decoder block are gated by the language-guided enhancement map produced by LTEM and then passed into the next decoder block.

#### 3.2.1 Target-sensitive Semantic Distance Module (TSDM)

Our TSDM enables segmentation-tailored contrastive alignment to bridge the pattern gaps between image and text features. Conventional image-text contrastive alignment defines positive pairs according to instance-level correspondence, i.e., whether the image and text come from the same patient, while all other samples are treated as negatives. This formulation implicitly assumes that only the paired image and text have identical semantics, whereas non-paired samples are semantically unrelated. However, this assumption is sub-optimal for medical image segmentation, as the semantic relationship between medical images and reports is inherently multi-granular (Wang et al., 2022a). For example, images and text from different patients may still describe highly similar target regions, while images and reports from the same disease category may differ substantially in ROI location, size, and shape. Therefore, our TSDM reformulates the pairing criterion in contrastive learning from “same patient” to “similar target semantics”, allowing image-text samples from different patients to be softly aligned if they correspond to similar segmentation targets.

As segmentation masks clearly delineate the target-related image regions that contain critical semantics, the semantic distance between different patients can be measured based on the intersections of their corresponding segmentation masks. Specifically, in the construction of contrastive alignment goals, the semantic distance $d_{p,q}$ for the $p$-th image and the $q$-th text in the batch is defined by the Dice coefficient between their corresponding ground truth segmentation masks. SoftMax is then applied to normalize $d_{p,q}$ across the batch dimension with a temperature $\tau_1$ (set as 10.0). Similarly, the current cosine distance $s_{p,q}$ between the $p$-th image features and the $q$-th text features in the batch is calculated and normalized using SoftMax with a temperature $\tau_2$ (set as 5.0).

A target-sensitive semantic contrastive loss is employed as the objective function of TSDM, calculated as a cross-entropy loss between the current distance $s_{p,q}$ and the target semantic distance $d_{p,q}$:

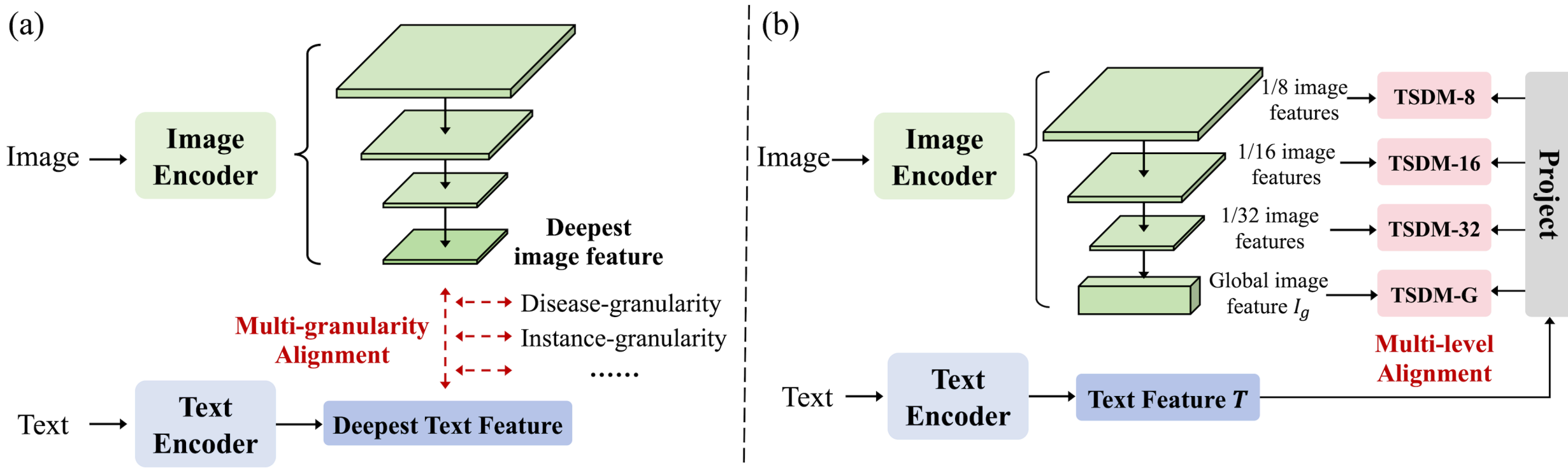

Figure 3. Illustration of (a) typical multi-granularity contrastive alignment and (b) our multi-level contrastive alignment strategy (MCAS).

$$L = -\frac{1}{Bz}\sum\sum d_{p,g} * \log(s_{p,q}), \tag{1}$$

where $Bz$ is the batch size. In our implementation, we computed the semantic contrastive loss in both image-to-text and text-to-image directions and then averaged them to derive a bi-directional contrastive loss.

### 3.2.2 Multi-level Contrastive Alignment Strategy (MCAS)

Our MCAS performs contrastive alignments at multiple feature pyramid levels, aiming to deliver fine-grained textual guidance to both high-level semantic representations and low-level localization-sensitive image features. Existing multi-granularity contrastive alignment methods (e.g., Wang et al., 2022a) typically define different semantic granularities, such as disease-granularity or instance-granularity correspondences, on the deepest image and text representations, as exemplified in Figure 3(a). Different from these approaches, our MCAS directly applies contrastive alignment to multi-level image features within the segmentation decoder. In this way, textual semantics can provide direct cross-modal supervision at several semantic resolutions, rather than relying solely on the deepest features to propagate alignment information to shallower layers.

As illustrated in Figure 3(b), our MCAS delivers fine-grained textual guidance on multi-scale image features containing rich low-level image details as well as global image features $I_g$ containing high-level image semantics. Specifically, the contrastive alignments conducted before three decoder blocks are denoted as TSDM-8/16/32, where the text features $T$ are semantically aligned with the image features in 1/8, 1/16, and 1/32 scales, respectively. Also, a global contrastive alignment, denoted as TSDM-G, is applied to align the global semantics between the global image features $I_g$ and the text features $T$. The text features $T$ are projected to match the dimension of image features at each level through linear projection. The final loss for multi-level contrastive alignments is defined as the average of bi-directional contrastive losses across all four levels.

### 3.2.3 Language-guided Target Enhancement Module (LTEM)

Our LTEM leverages the aligned image-text features to highlight critical image regions. Although LTEM is conceptually related to attention-based feature reweighting, its underlying mechanism differs from standard cross-attention or conventional spatial attention. Standard cross-attention typically performs direct feature interaction between image and text features at the same stage, while common spatial attention usually learns an attention map from visual features alone without explicit cross-modal semantic calibration. In contrast, our LTEM is designed as a top-down target enhancement mechanism built upon the semantically aligned image-text representations produced by TSDM and MCAS. Its purpose is to transfer target-aware textual guidance from deeper, text-aware representations to shallower decoder features, which contain richer spatial details and are more responsible for accurate target localization and boundary delineation.

Since the deeper layers of CNNs tend to contain high-level semantic information, while the shallower layers tend to contain low-level detail information (Selvaraju et al., 2017), we calculate a language-guided cross-attention map based on high-level image-text features to gate the low-level image features output from decoder blocks. Specifically, we calculate the normalized dot product similarity between the $i$-th image sub-region features $I_i$ and the global text features $T_g$ (i.e., globally pooling $T$ across the word length), which is defined as:

$$w_i = I_i^T T_g. \tag{2}$$

Then, we normalize $w_i$ across the image sub-region dimension using SoftMax with a temperature $\tau_3$ (set as 5.0). The normalized map is reshaped and upsampled by a factor of 2, resulting in a language-guided enhancement map. Finally, the enhancement map is multiplied by the decoded image features, which is then passed into the next decoder block or the final segmentation head.

## 4. Experimental Setup

### 4.1. Datasets and Preprocessing

We evaluated our method on four well-established public benchmark datasets, QaTa-COV19 (Degerli et al., 2022), MosMedData (Morozov et al., 2020), Kvasir-SEG (Jha et al., 2019), and Bkai-polyp (Phan et al., 2021), involving X-ray, computed tomography (CT), and endoscopic images paired with textual reports.

The QaTa-COV19 dataset contains 9258 COVID-19 Chest X-rays with ground-truth segmentation labels for the COVID-19 infected regions. The official training set of the QaTa-COV19 dataset contains 7145 samples, while the test set contains 2113 samples. Following the preprocessing procedure in Zhong et al.'s work (2023), the official training set was split into train and validation sets with a distribution of 80% and 20%. The MosMedData dataset contains 2729 CT slices depicting pulmonary infections with ground truth segmentation labels. We followed the preprocessing procedure in Li et al. (2023)'s work, splitting the dataset into 2183, 273, and 273 samples for training, validation, and testing. For language-guided segmentation, Li et al. (2023) extended the QaTa-COV19 and MosMedData datasets with matched textual reports. Each report contains three sentences: the first sentence indicates the presence of infection, the second outlines the number of involved regions, and the third describes their location, which provide rich semantics that can be used to guide the lesion segmentation.

The Kvasir-SEG and Bkai-polyp datasets, respectively, contain 1000 endoscopic images with ground-truth segmentation labels of gastrointestinal polyps. We followed the preprocessing procedure in Poudel et al. (2023)'s work, splitting the datasets with a ratio of 8:1:1 for training, validation, and testing. Poudel et al. (2023) also extended these two datasets with matched textual reports, describing the polyp size, number, color, and location in a free-text format.

### 4.2. Implementation Details

The deep learning models were trained using a Nvidia 24 GB RTX3090 GPU with a batch size of 32. PyTorch was employed with PyTorch Lightning as the wrapper for training and testing. The AdamW optimizer was used with a learning rate set to 3e-4, and a cosine annealing scheduler was employed with the minimal learning rate set to 1e-6. All images were resized to a dimension of 224×224. Image augmentation was applied by a random zoom with a 10% probability.

### 4.3. Experimental Settings

Our TMCA was extensively compared to existing medical image segmentation methods: (i) image-only segmentation methods including U-Net (Ronneberger et al., 2015), U-Net++ (Zhou et al., 2018), Attention U-Net (Oktay et al., 2018), and nnU-Net (Isensee et al., 2021); (ii) vision-language pretraining (VLP)-based methods including CLIP (Radford et al., 2021) and GloRIA (Huang, S. et al., 2021); and (iii) language-guided segmentation methods including LViT (Li et al., 2023), TGANet (Tomar et al., 2022), Ariadne (Zhong et al., 2023), RecLMIS (Huang et al., 2025), EviVLM (Pan et al., 2026), and HCFNet (Zhou et al., 2025). Note that CLIP and GloRIA were not evaluated on endoscopic polyp datasets (Kvasir-SEG and Bkai-polyp) as they are not pretrained on this domain.

Moreover, ablation analyses were conducted to investigate the individual effectiveness of the three proposed components (TSDM, MCAS, and LTEM). A baseline segmentation model was built for comparison in the ablation studies, which has the same encoder-decoder architecture as ours but removes all the components related to our target-informed multi-level contrastive alignments. We also evaluated the performance under textual input truncation, where the report is progressively truncated to different token lengths.

We adopted two commonly used segmentation evaluation metrics: Dice Similarity Coefficient (Dice) and mean Intersection over Union (mIoU), two widely recognized metrics for measuring spatial overlap between the predicted and ground truth segmentation masks. In general, a higher Dice or a higher mIoU indicates a better segmentation performance.

## 5. Results

### 5.1. Comparison with Existing Methods

Table 1 presents a quantitative comparison with existing methods for medical image segmentation. Language-guided methods achieved better overall performance than image-only and VLP-based methods, while VLP-based methods failed to outperform image-only methods despite using textual information during pretraining. Among language-guided methods, HCFNet outperformed other recently proposed methods (RecLMIS and EviVLM), achieving the second-best performance. Nevertheless, our TMCA still outperformed HCFNet, especially by a large margin on the Kvasir-SEG and Bkai-polyp datasets.

Moreover, most comparison methods exhibited inconsistent performance across four datasets. Among image-only methods, U-Net++ achieved the best performance on the QaTa-COV19 and Bkai-polyp datasets, while nnU-Net and Attention U-Net achieved the best performance on the MosMedData and Kvasir-SEG datasets, respectively. For language-guided methods, Ariadne achieved the second-best Dice result on the Bkai-polyp dataset, but its results on the other datasets are worse than HCFNet. Similarly, EviVLM achieved the second-best mIoU result on the MosMedData dataset, but its results on the QaTa-COV19 and Bkai-polyp datasets are even worse than Ariadne. In contrast, our TMCA achieved the best performance consistently across all evaluation datasets.

Table 1. Quantitative comparison between our TMCA and existing methods for medical image segmentation.

| Method | QaTa-COV19 | | MosMedData | | Kvasir-SEG | | Bkai-polyp | |
|---|---|---|---|---|---|---|---|---|
| | mIoU (%) | Dice (%) | mIoU (%) | Dice (%) | mIoU (%) | Dice (%) | mIoU (%) | Dice (%) |
| ***Image-only*** | | | | | | | | |
| U-Net | 70.92 | 82.99 | 50.73 | 64.60 | 69.62 | 82.09 | 79.06 | 88.31 |
| U-Net++ | 71.96 | 83.69 | 58.39 | 71.75 | 69.58 | 82.06 | 79.87 | 88.81 |
| nnU-Net | 70.81 | 80.42 | 60.36 | 72.59 | 75.48 | 85.64 | 77.14 | 87.36 |
| Attention U-Net | 70.06 | 82.40 | 52.82 | 66.34 | 74.19 | 85.18 | 72.97 | 84.26 |
| ***VLP-based*** | | | | | | | | |
| CLIP | 70.66 | 79.81 | 59.64 | 71.97 | - | - | - | - |
| GloRIA | 70.68 | 79.94 | 60.18 | 72.42 | - | - | - | - |
| ***Language-guided*** | | | | | | | | |
| LViT | 73.79 | 84.92 | 61.33 | 74.57 | 77.04 | 87.03 | 75.19 | 85.84 |
| TGANet | 70.75 | 79.87 | 59.28 | 71.81 | 83.30 | 89.82 | 84.09 | 90.23 |
| Ariadne | 81.45 | 89.78 | 63.04 | 77.33 | 80.70 | 89.32 | 88.02 | <u>93.63</u> |
| RecLMIS | 77.00 | 85.22 | 65.07 | 77.48 | 78.76 | 85.75 | 78.42 | 88.63 |
| EviVLM | 77.34 | 85.79 | <u>65.81</u> | 77.64 | 82.45 | 89.63 | 85.25 | 90.76 |
| HCFNet | <u>83.06</u> | <u>90.75</u> | 65.75 | <u>79.34</u> | <u>83.49</u> | <u>90.18</u> | <u>88.10</u> | 93.14 |
| TMCA (Ours) | **83.27** | **90.87** | **65.94** | **79.56** | **85.54** | **92.20** | **89.91** | **94.69** |

**Bold**/<u>Underline</u>: The best/second-best result in each column.

## 5.2. Ablation Analysis of Each Component

Table 2 presents our first ablation study, where our three technical contributions (MCAS, LTEM, and TSDM) were individually added to the baseline segmentation model or removed from the complete TMCA. Starting from the baseline model, adding each component individually improves the segmentation performance, indicating that all three modules contribute positively to TMCA. Among them, MCAS yielded the largest individual gain, which demonstrates the importance of performing contrastive alignment at multiple feature levels for language-guided segmentation. TSDM also brought clear improvements by introducing target-sensitive semantic distances, while LTEM further enhances target-related visual regions through aligned textual guidance. When combining components, the performance continued to increase, with TSDM+MCAS achieving the second-best results among all variants. This suggests that target-sensitive distance modeling and multi-level alignment are highly complementary. Finally, the complete TMCA equipped with all three components achieved the best performance, reaching 85.54% mIoU, 92.20% Dice, and 97.82% accuracy, outperforming the baseline by 4.84%, 2.88%, and 0.70%, respectively.

Table 2. Ablation results of incrementally adding each component on the Kvasir-SEG dataset.

| TSDM | LTEM | MCAS | mIoU (%) | Dice (%) | Accuracy (%) |
|---|---|---|---|---|---|
| × | × | × | 80.70 | 89.32 | 97.12 |
| √ | × | × | 82.07 | 90.12 | 97.25 |
| × | √ | × | 81.21 | 89.55 | 97.18 |
| × | × | √ | 83.41 | 90.96 | 97.46 |
| × | √ | √ | 84.58 | 91.64 | 97.65 |
| √ | × | √ | <u>84.74</u> | <u>91.73</u> | <u>97.67</u> |
| √ | √ | × | 82.30 | 90.23 | 97.29 |
| √ | √ | √ | **85.54** | **92.20** | **97.82** |

**Bold**/<u>Underline</u>: The best/second-best result in each column.

## 5.3. Ablation Analysis on Target-sensitive Semantic Distance

Table 3 presents the ablation analysis on different semantic distance definitions for contrastive alignment, where the proposed TSDM was compared to three alternative strategies: instance-level pairing, disease label-based pairing, and text feature similarity-based semantic distance. Instance-level pairing follows the common contrastive learning setting, where only the matched image-text pair from the same sample is treated as a positive pair. Disease label-based pairing extracts ROI-related keyword and location information from the textual reports as classification labels, and image-text samples with the same label are treated as semantically paired. Text feature similarity-based semantic distance directly calculates the similarity between text features and uses it as the semantic distance between image-text samples. As shown in Table 3, TSDM achieved the best performance, and Disease label-based pairing obtained the second-best results. Text feature similarity-based semantic distance achieved the lowest performance.

Table 3. Ablation analysis on target-sensitive semantic distance on the Kvasir-SEG dataset.

| Method | mIoU (%) | Dice (%) | Accuracy (%) |
|---|---|---|---|
| Instance-level Pairing | 84.58 | 91.64 | 97.65 |
| Disease label-based | <u>85.15</u> | <u>91.93</u> | <u>97.73</u> |
| Text feature similarity-based | 83.49 | 90.86 | 97.45 |
| TSDM (Ours) | **85.54** | **92.20** | **97.82** |

**Bold**/<u>Underline</u>: The best/second-best result in each column.

### 5.4. Ablation Analysis on Multi-level Contrastive Alignments

Table 4 presents the ablation analysis on Multi-level Contrastive Alignments, where different levels of target-informed contrastive alignments were added to the baseline segmentation model. We noticed that adding TSDM-G to align global image-text semantics has yielded an increase of 1.60/0.91% in mIoU/Dice over the baseline model, while adding TSDM at 1/32, 1/16, and 1/8 scales (TSDM-32/16/8) further improved segmentation performance, eventually attaining an increase of 4.84/2.88% in mIoU/Dice.

Table 4. Ablation results of multi-level contrastive alignments on the Kvasir-SEG dataset.

| Method | mIoU (%) | Dice (%) | Accuracy (%) |
|---|---|---|---|
| Baseline | 80.70 | 89.32 | 97.12 |
| + TSDM-G | 82.30 | 90.23 | 97.29 |
| + TSDM-G/32 | 83.74 | 91.14 | 97.55 |
| + TSDM-G/32/16 | <u>84.35</u> | <u>91.51</u> | <u>97.63</u> |
| + TSDM-G/32/16/8 | **85.54** | **92.20** | **97.82** |

**Bold**/<u>Underline</u>: The best/second-best result in each column.

### 5.5. Ablation Analysis on Language-guided Target Enhancement

Table 5 presents the ablation analysis on language-guided target enhancement, where the proposed LTEM was compared to three variants: removing LTEM, replacing LTEM with a standard cross-attention module, and replacing LTEM with a spatial attention module. The cross-attention variant uses standard image-text cross-attention to further enhance visual features after decoder blocks, while the spatial attention variant learns visual attention merely based on image features for feature reweighting. As shown in Table 5, the complete model with LTEM achieved the best performance. Replacing LTEM with cross-attention improved the performance over removing LTEM, while replacing it with spatial attention provided only marginal improvement.

Table 5. Ablation analysis on language-guided target enhancement on the Kvasir-SEG dataset.

| Method | mIoU (%) | Dice (%) | Accuracy (%) |
|---|---|---|---|
| Remove LTEM | 84.74 | 91.73 | 97.67 |
| Replaced by Cross-attention | <u>85.08</u> | <u>91.90</u> | <u>97.72</u> |
| Replaced by Spatial Attention | 84.78 | 91.75 | 97.68 |
| With LTEM (Ours) | **85.54** | **92.20** | **97.82** |

**Bold**/<u>Underline</u>: The best/second-best result in each column.

### 5.6. Robustness Analysis under Text Input Truncation

Table 6 presents the robustness analysis under textual input truncation. In our implementation, the text encoder uses max_length=24 with padding='max_length' by default, while the average report length in the evaluation datasets is approximately 11.5–12.5 tokens. For the Kvasir-SEG dataset, the first 4–6 tokens usually describe the number, shape, and color of the lesion, whereas the tokens around 6–12 often contain key spatial localization descriptions. Therefore, truncating the input reports to 16 or 12 tokens mainly removes redundant padding tokens and preserves the major target-related semantics, leading to slightly improved performance compared with the original text input. When the textual input is further truncated to 8 or 4 tokens, the segmentation performance decreases to 83.54%/90.95% and 81.13%/89.52% in mIoU/Dice, respectively. This degradation is reasonable because excessive truncation may remove key location cues required for accurate localization-oriented image-text fusion. In addition, the no-text setting shows a clear performance drop, achieving only 77.22% mIoU and 87.13% Dice, which further confirms the importance of textual guidance for medical image segmentation.

Table 6. Robustness analysis under textual input truncation on the Kvasir-SEG dataset.

| Method | mIoU (%) | Dice (%) | Accuracy (%) |
|---|---|---|---|
| No text | 77.22 | 87.13 | 96.58 |
| Truncated to 4 tokens | 81.13 | 89.52 | 97.19 |
| Truncated to 8 tokens | 83.54 | 90.95 | 97.50 |
| Truncated to 12 tokens | **85.55** | **92.22** | **97.83** |
| Truncated to 16 tokens | <u>85.54</u> | <u>92.21</u> | <u>97.82</u> |
| Original text (Ours) | <u>85.54</u> | 92.20 | <u>97.82</u> |

**Bold**/<u>Underline</u>: The best/second-best result in each column.

### 5.7. Visualization Analysis

Figure 4 shows a qualitative comparison with existing methods for medical image segmentation, demonstrating that our TMCA accurately segmented lesions even in cases of low image contrast and image quality (rows 1, 2, 3, and 5). Other methods, especially image-only methods (U-Net and U-Net++), tended to produce false positive segmentation results (rows 1-7) and/or failed to segment lesions with low contrast to the background (rows 2,5, and 6). Other language-guided methods (LViT and Ariadne) used textual guidance to reduce false positive results (e.g., row 1, where the text specified that there is no infected region in left lung) and to segment challenging regions (e.g., row 6, where the text indicates that there is polyp at the top of the image). Nevertheless, our TMCA further refined the textual guidance through the proposed TSDM and applied it to subtle image details through the proposed MCAS and LTEM, achieving the most accurate segmentation results across all four datasets.

Figure 5 presents a visualization analysis of image-text alignment. In each case, the textual report describes specific target-related semantics, including lesion number, appearance, and spatial location, while the corresponding objects in the ground truth and segmentation results are marked with the same colors. Red boxes denote erroneous predictions, including false positives, false negatives, or regions inconsistent with the textual description. As shown in Figure 5, the baseline model tends to produce predictions that are poorly aligned with the text, such as missing text-indicated regions, segmenting irrelevant areas, or failing to distinguish multiple targets with different locations. In contrast, our TMCA generates segmentation results that are more consistent with both the ground truth and the textual descriptions. This improvement is particularly evident in cases involving multiple infected regions or specific polyp locations, where TMCA better preserves the correspondence between the language-described targets and the predicted masks. These visualization results provide intuitive evidence that the proposed target-informed multi-level contrastive alignments improve image-text semantic alignment and enable more accurate language-guided segmentation.

## 6. Discussion

Through extensive experimental validation, our study reveals three main findings: (i) Our TMCA outperformed state-of-the-art medical image segmentation methods and exhibited great generalizability across various imaging modalities and segmentation tasks; (ii) The proposed TSDM, LTEM, and MCAS collectively enhanced the segmentation performance by enabling finer-grained textual semantic guidance on critical image regions; (iii) Compared to the widely used single-level contrastive alignment, imposing multi-level contrastive alignments on both global and local image features contributed to improved segmentation performance.

In the quantitative comparison between our TMCA and state-of-the-art methods (Table 1), language-guided methods leveraging textual guidance, such as Ariadne, EviVLM, and HCFNet, consistently outperformed image-only segmentation methods such as U-Net and U-Net++. We attributed this to the integration of additional textual information that provided location and size related to ROI targets, as exemplified in Figure 4. The results of LViT are relatively lower among language-guided methods, and this is likely due to LViT only employing a simple tokenization layer to extract text features, which limits the ability to capture high-level textual semantics related to segmentation. Ariadne employed a well-pretrained text encoder capable of extracting rich, high-level textual semantics, thereby achieving improved segmentation performance. HCFNet outperformed all other language-guided methods, and we attribute this partly to the use of contrastive learning for aligning image-text patterns. Nevertheless, our TMCA still outperformed HCFNet consistently across all evaluation datasets by introducing target-informed multi-level contrastive alignments to bridge the image-text pattern gaps in a segmentation-tailored manner. With reduced image-text pattern gaps, textual information can be more effectively incorporated for segmentation, as evidenced in Figure 4 showing that our TMCA more precisely employed the textual semantics to guide segmentation in hard negative cases.

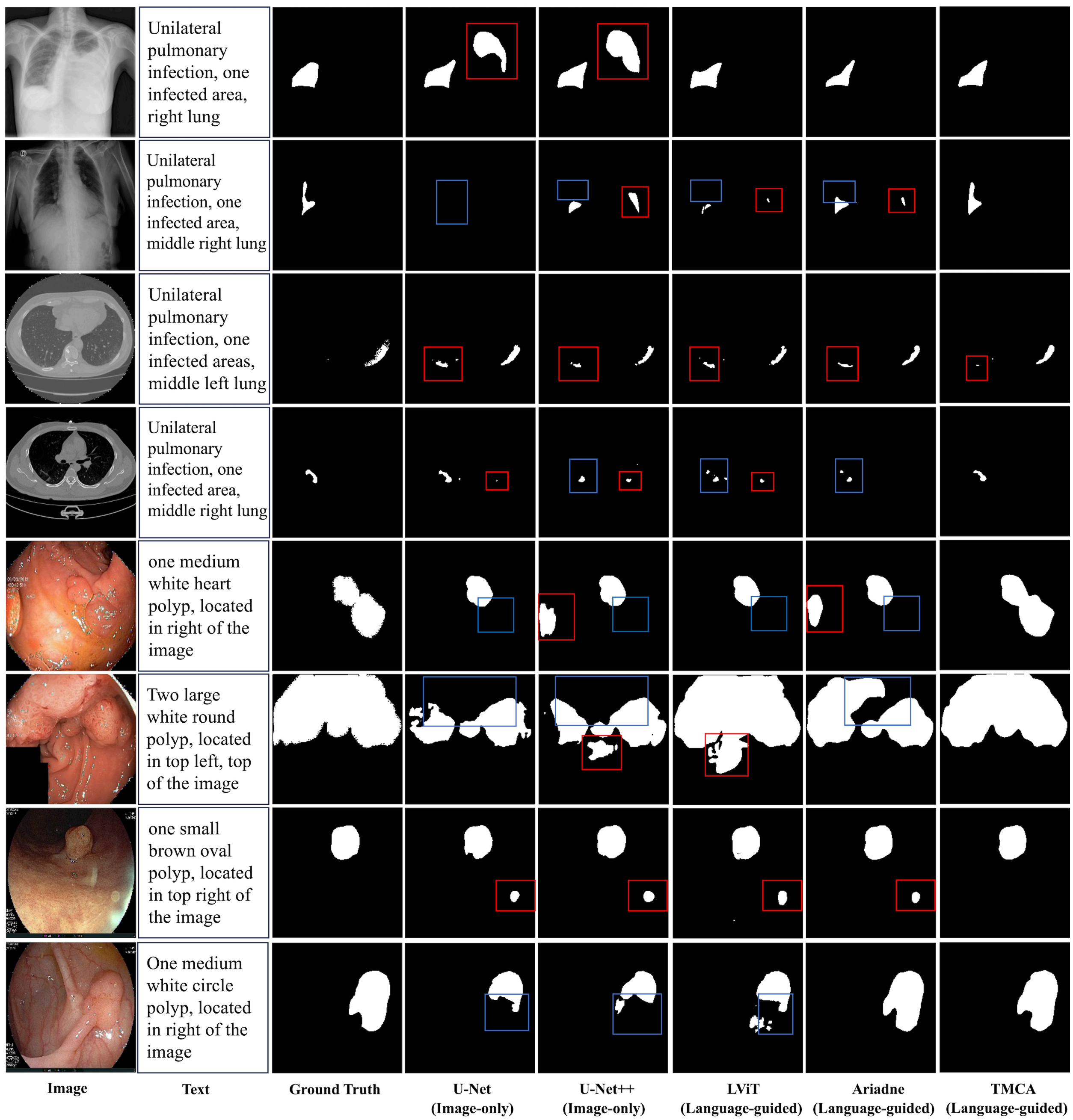

Figure 4. Qualitative comparison with existing methods for medical image segmentation on the QaTa-COV19 (rows 1-2), MosMedData (rows 3-4), Kvasir-SEG (rows 5-6), and Bkai-polyp (rows 7-8) datasets. Red boxes indicate the regions falsely identified (false positive) and blue boxes indicate the regions not identified (false negative).

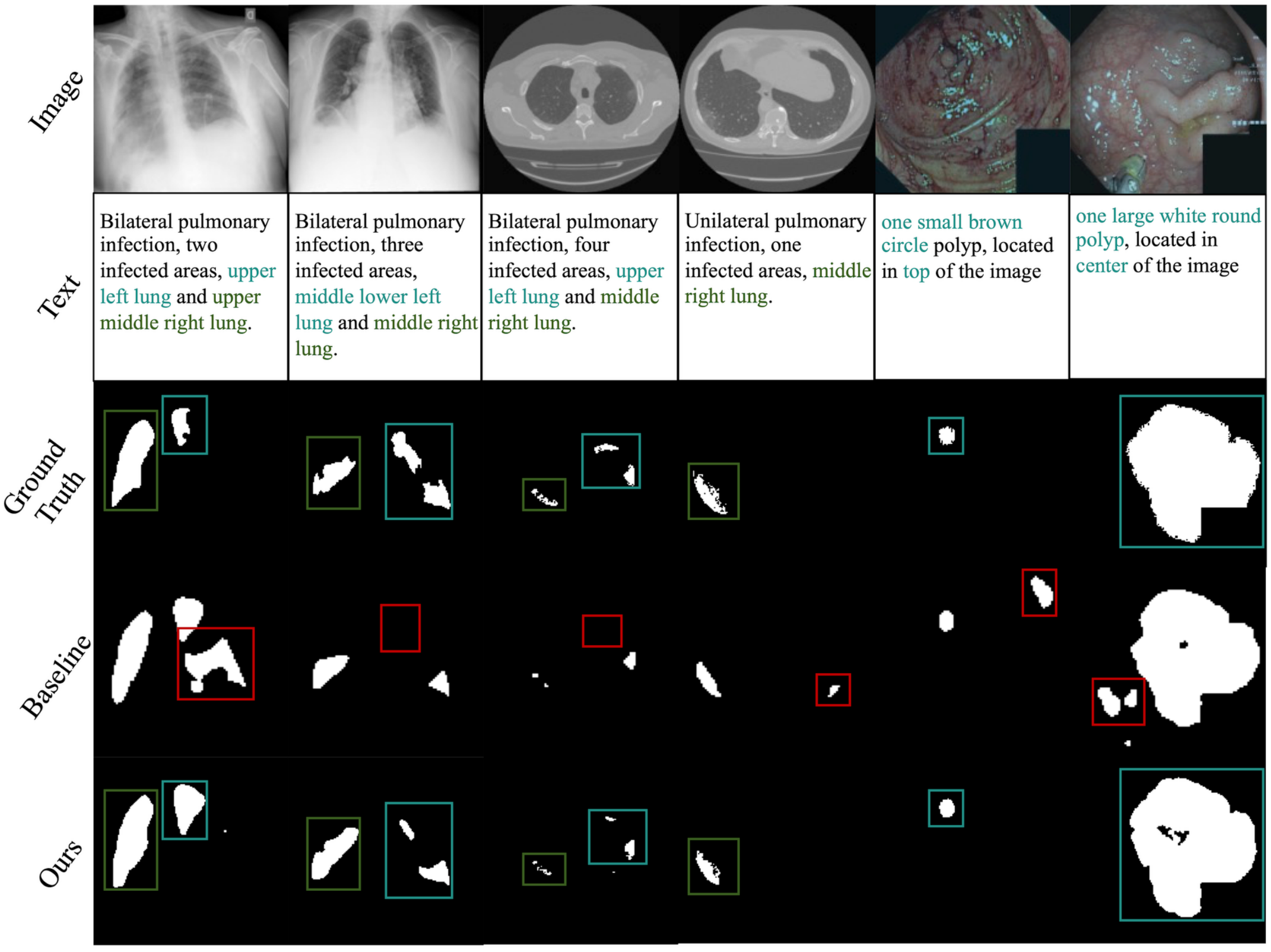

Figure 5. Visualization analysis on image-text alignment. The corresponding objects are marked in the same color, while red boxes indicate errors.

In the ablation analysis of each technical component (Table 2), employing standard contrastive alignments with MCAS has already yielded superior segmentation performance over the baseline model, while adding LTEM and TSDM further enhanced the performance. These results demonstrate that Dice-based target-sensitive semantic distance is especially suitable for segmentation-oriented multimodal alignment because it evaluates similarity on effective target regions rather than on global image content. This suppresses the influence of large homogeneous background areas and yields a target-aware estimate of cross-modal consistency. Meanwhile, multi-level alignment is beneficial because hierarchical visual features carry complementary information: shallow layers preserve boundary and localization cues, whereas deeper layers encode more abstract semantics. Supervising multiple levels therefore reduces cross-modal discrepancies more effectively than relying on a single deepest representation. Finally, LTEM improves target localization by transferring semantically aligned textual guidance from deeper layer features to shallow visual layers, allowing segmentation-sensitive features to focus more accurately on clinically relevant regions.

In the ablation analysis on target-sensitive semantic distance (Table 3), the results indicate that the choice of semantic distance has a clear influence on contrastive alignment for language-guided segmentation. Text feature similarity-based semantic distance performs even worse than the common instance-level pairing strategy. A possible reason is that the textual reports in the current datasets are highly structured and share similar sentence patterns, causing most text features to exhibit high similarity. As a result, the truly discriminative target-related information can be overwhelmed by general textual similarity, making it difficult to distinguish different ROI semantics. Disease label-based pairing outperforms the common instance-level pairing strategy because it directly uses key semantic information, such as ROI-related attributes and locations, to define positive pairs. However, it still underperforms our Dice-based TSDM, because classification labels provide only coarse semantic grouping and cannot fully characterize target-level differences in ROI size, shape, and spatial overlap. In contrast, TSDM measures semantic relatedness directly based on segmentation masks, enabling more fine-grained target-sensitive alignment that is better suited for medical image segmentation.

In the ablation analysis on multi-level contrastive alignments (Table 4), our TMCA yielded larger improvements when more levels of contrastive alignments were applied. We suggest that this is because, with contrastive alignments applied at higher-scale image features, finer-grained textual guidance is integrated with more detailed image contexts for accurate segmentation. When only the global text and image features at the deepest layer are aligned via contrastive alignment, information integration at shallower layers may become confused and distract the decoder's features from important regions. These shallow layers, which focus on local image details such as the ROI boundaries, are heavily reliant on high-level image semantics and auxiliary textual guidance to identify attentive regions. By incorporating contrastive alignments at multiple pyramid levels, our TMCA can leverage textual semantics to directly guide the modeling of local image details. This approach also enables our LTEM, which associates each image sub-region with text by calculating the feature similarities, to provide more accurate localization guidance at multiple pyramid levels. This is particularly important when the ROI targets are small and primarily rely on subtle image details to identify. Intuitively, this observation is also consistent with deep supervision, where additional auxiliary supervision provides supplementary guidance on the intermediate modeling of neural networks (Le'Clerc Arrastia et al., 2021; Zhao et al., 2021).

In the ablation analysis on language-guided target enhancement (Table 5), the results demonstrate that the proposed LTEM is more effective than generic attention-based feature enhancement. Standard cross-attention mainly performs same-level feature interaction and does not explicitly transfer semantically aligned target guidance from deeper representations to shallower localization-sensitive features. Spatial attention provides only marginal improvement over removing LTEM, suggesting that visual-only attention is insufficient to reliably identify target-related regions without textual semantic calibration. In contrast, LTEM leverages semantically aligned image-text representations as top-down target priors to guide low-level decoder features, enabling more precise enhancement of clinically relevant regions and better suppression of irrelevant background responses.

In addition, we observed that the improved performance of TMCA is achieved with only moderate additional overhead. Compared with the baseline model, TMCA slightly increases the total parameter number from 153.581M to 155.234M and the training memory from 5.585 GB to 6.558 GB, while the inference time increases only from 12.097 ms to 13.913 ms. Although the FLOPs increase from 22.973G to 30.170G due to the additional alignment and enhancement operations, the actual inference-time overhead remains small. This is because TSDM and MCAS introduce contrastive supervision merely during training and do not require extra computation during inference. During inference, the main overhead comes from LTEM, which performs lightweight language-guided target enhancement. Consequently, TMCA improves segmentation accuracy through target-informed multi-level contrastive alignment while maintaining practical computational efficiency.

Language-guided medical image segmentation has proven useful in practical scenarios where clinicians can provide interactive language guidance (Li et al., 2023) and also in retrospective studies where image-text pairs are readily available, as in this study. Nevertheless, a key limitation is that existing works relied on the text inputs that were carefully crafted by humans to be concise and direct for describing ROIs (e.g., "one small brown oval polyp, located in the top right of the image"). In routine clinical workflows, however, medical reports can be much longer, including important information such as patient history, scanning protocols, relationships to adjacent organs, and details about the lack of involvement of other organs/structures. Such reports pose challenges to language-guided segmentation methods as their textual semantics are implicit and can be difficult to extract. This may be mitigated by recent natural language processing (NLP) techniques; for example, lengthy textual reports can be refined and structured using large language models (LLMs) (Touvron et al., 2023; Wu et al., 2024). As future work, we will investigate the use of LLM to process such clinical imaging reports and evaluate its performance with our TMCA, and also will curate segmentation datasets containing more complex clinical reports. Another limitation is that the current study does not include cross-dataset evaluation. Similar to most existing language-guided medical segmentation studies, our method is trained and evaluated under dataset-specific image-text annotation settings, because different datasets may involve different imaging modalities, anatomical targets, segmentation definitions, and text annotation styles. Therefore, direct training on one dataset and testing on another is not straightforward and may be difficult to interpret. In future work, we will further investigate the cross-dataset generalization of TMCA under heterogeneous image-text annotation settings. Also, we will explore language-guided segmentation for 3D medical images, which is more complicated and requires finer-grained textual guidance.

## 7. Conclusion

In this study, we have outlined a Target-informed Multi-level Contrastive Alignment framework (TMCA) for language-guided medical image segmentation, which bridges the inherent image-text pattern gaps to optimize cross-modal information integration through multi-level fine-grained target-sensitive contrastive alignments. Our TMCA introduces a Target-sensitive Semantic Distance Module (TSDM) and a Language-guided Target Enhancement Module (LTEM) to deliver finer-grained textual guidance for critical image regions, and also introduces a Multi-level Contrastive Alignment Strategy (MCAS) to direct textual guidance to multi-scale image features with rich image details. Extensive experiments demonstrated that our TMCA consistently outperformed state-of-the-art medical image segmentation methods across four well-established benchmark datasets and thus exhibited good generalizability across various segmentation tasks and medical imaging modalities.

## Acknowledgement

This work was supported by the Australian Research Council (ARC) under Grant DP200103748 and the Zhongguancun Academy under Grant XTS0071.